# CRYSTAL: Inducing a Conceptual Dictionary *

Stephen Soderland, David Fisher,
Jonathan Aseltine, Wendy Lehnert
Department of Computer Science
University of Massachusetts, Amherst, MA 01003-4610
{soderlan dfisher aseltine lehnert}@cs.umass.edu

## Abstract

One of the central knowledge sources of an information extraction (IE) system is a dictionary of linguistic patterns that can be used to identify references to relevant information in a text. Automatic creation of conceptual dictionaries is important for portability and scalability of an IE system. This paper describes CRYSTAL, a system which automatically induces a dictionary of "concept-node definitions" sufficient to identify relevant information from a training corpus. Each of these concept-node definitions is generalized as far as possible without producing errors, so that a minimum number of dictionary entries cover the positive training instances. Because it tests the accuracy of each proposed definition, CRYSTAL can often surpass human intuitions in creating reliable extraction rules.

## 1 Information Extraction

An information extraction (IE) system analyzes unrestricted natural language text and produces a representation of the information from the text which is considered relevant to a particular application. Information extracted from the text is represented as case frames, called "concept nodes" (CN's) in the University of Massachusetts BADGER sentence analyzer, which performs selective concept extraction similar to that of the previous CIRCUS system [Lehnert, 1991; Lehnert et al., 1993].

The domain knowledge needed to identify relevant references is stored in a dictionary of "CN definitions" that describe the local syntactic and semantic context in which relevant information is likely to be found. Before the CN definition is applied, BADGER segments the input text to identify syntactic constituents such as subject, verb phrase, direct and indirect object, and prepositional phrases, and also looks up the semantic class of each word in a domain-specific semantic lexicon. A CN definition specifies a set of syntactic and semantic constraints that must be satisfied for the definition to apply to a segment of text.

Examples in this paper are from a medical domain, where the task is to analyze hospital discharge reports and identify references to "diagnosis" and to "sign or symptom". These are further classified with subtypes.

| Diagnosis: | Sign or Symptom: |
|---|---|
| confirmed | present |
| ruled out | absent |
| suspected | presumed |
| pre-existing | unknown |
| past | history |

The example shown in Figure 1 is a CN definition from this domain that identifies references to absent symptoms. This CN definition extracts the phrase in the direct object buffer when the subject buffer has the word "patient" (semantic class <Patient or Disabled Group>), the verb is "denies" in the active voice, and the direct object has the semantic class <Sign or Symptom>.

```
CN-type: Sign or Symptom
Subtype: Absent
Extract from Direct Object
Active voice verb
Subject constraints:
    words include  "PATIENT"
    head class:    <Patient or Disabled Group>
Verb constraints:
    words include  "DENIES"
Direct Object constraints:
    head class     <Sign or Symptom>
```

Figure 1: A CN definition to identify "sign or symptom, absent"

This CN definition would extract "any episodes of nausea" from the sentence "The patient denies any episodes of nausea". It would fail to apply to the sentence "Patient denies a history of asthma", since asthma is of semantic class <Disease or Syndrome>, which is not a subclass of <Sign or Symptom>. For the hospital discharge report domain, we are using a semantic lexicon

*This research was supported by NSF Grant no. EEC-9209623, State/ Industry/ University Cooperative Research on Intelligent Information Retrieval. Thanks also to David Aronow for help as our medical domain expert.

and semantic hierarchy derived from the Unified Medical Language Systems (UMLS) medical MetaThesaurus and Semantic Network [Lindberg et al., 1993], which is currently under development by the National Library of Medicine.

A dictionary of CN definitions for this domain is specific to the semantics and writing style of hospital discharge records and could not be transferred to other applications. A new conceptual dictionary must be constructed for each IE application. The definitions must be general enough to cover previously unseen instances, but at the same time constrained tightly enough to avoid over generalizing to instances that do not contain the desired information.

A tool that automatically generates a dictionary of CN definitions is needed to ensure that BADGER can be easily ported to new domains. The CRYSTAL dictionary induction tool is one of the first systems to automatically create a conceptual dictionary from a training corpus. Before presenting CRYSTAL, we will describe in more detail the CN definitions used by the BADGER sentence analyzer, which CRYSTAL must generate automatically.

## 2 Concept Node Definitions

A concept node (CN) is a case frame instantiated by the BADGER sentence analyzer to represent relevant information identified in the text. The CN has two fixed slots, CN type and subtype, as well as slots to hold extracted information, which are usually noun phrases from the text. A CN is instantiated from a text segment when the constraints of a CN definition are satisfied.

These constraints operate on major syntactic constituents: the subject, verb phrase, direct or indirect object, and prepositional phrases. Any of these constituents may be tested for a sequence of specific words, for specific semantic classes in the head noun of a phrase, or for specific semantic classes in the modifiers of a phrase. The verb can be further constrained with respect to active or passive voice.

Figure 2 shows a CN definition that identifies pre-existing diagnoses with a set of constraints that could be summarized as "... was diagnosed with recurrence of (body part) (disease)". Here the information to be extracted is found in a prepositional phrase which must have the preposition "with" and contain the words "recurrence of". The prepositional phrase must have a head noun whose semantic class is a <Disease or Syndrome> and a modifying term whose class is a <Body Part or Organ>. This CN definition applies to a sentence such as "The patient was diagnosed with a recurrence of laryngeal cancer". Since there are no constraints on the subject, the text segment is free to have any subject, including a relative pronoun or an omitted subject.

Will this CN definition reliably identify only pre-existing diagnoses? Perhaps in some texts the recurrence of a disease is actually a principal diagnosis of the current hospitalization and should be identified as "diagnosis, confirmed" or may be a condition that no longer exists and should be identified as "diagnosis, past". In such cases the CN definition will produce an extraction error. On the other hand, this definition might be reli-

```
CN-type: Diagnosis
Subtype: Pre-existing
Extract from Prep. Phrase "WITH"
Passive voice verb
Verb constraints:
    words include   "DIAGNOSED"
Prep. Phrase constraints:
    preposition =   "WITH"
    words include   "RECURRENCE OF"
    modifier class  <Body Part or Organ>
    head class      <Disease or Syndrome>
```

Figure 2: A CN definition for "diagnosis, pre-existing"

able, but miss some valid examples that it would cover if the constraints were relaxed slightly.

Judgments about how tightly to constrain a CN definition are difficult to make *a priori*, and would require careful consideration by someone who combines domain expertise with a deep understanding of the BADGER sentence analyzer. An alternate to manually engineering these CN definitions is to induce them automatically from a training corpus of representative texts that have been annotated by a domain expert. This is the approach taken by CRYSTAL, which we will describe in the following section.

## 3 The CRYSTAL Dictionary Induction Tool

CRYSTAL derives a domain-specific dictionary of CN definitions from a training corpus, initializing the dictionary with a CN definition for each positive training instance. These initial CN definitions are designed to extract the relevant phrase in the training instance that motivated them, but are too specific to apply broadly to previously unseen sentences.

The main work of CRYSTAL is to gradually relax the constraints on these initial definitions to broaden their coverage, while merging similar definitions to form a more compact dictionary. The CN definitions in CRYSTAL's final dictionary are generalized as much as possible without producing extraction errors on the training corpus.

### 3.1 Creating Initial CN Definitions

The first step in dictionary creation is the annotation of a set of training texts by a domain expert. Each phrase that contains information to be extracted is bracketed with SGML-style tags to mark the appropriate CN type and subtype. A team of three nurses under the supervision of a physician annotated our training documents for us. The annotated texts are then segmented by the BADGER sentence analyzer to create a set of training instances. Each instance is a text segment, generally a simple clause, some of whose syntactic constituents may be tagged as positive instances of a particular CN type and subtype.

CRYSTAL begins its induction with a dictionary of CN definitions built from each instance that contains the CN type and subtype being learned. If a training instance has its subject buffer tagged as "diagnosis" with subtype "pre-existing", an initial CN definition is created that extracts the phrase in the subject buffer as a pre-existing diagnosis. The constraints on an initial CN definition are derived from the words and classes found in the motivating instance.

Before the induction process begins, CRYSTAL cannot predict which characteristics of an instance are essential to the CN definition and which are merely accidental features. CRYSTAL encodes all the details of the text segment as constraints to the initial CN definition, requiring the exact sequence of words and the exact sets of semantic classes in each syntactic buffer. CRYSTAL will later learn which constraints should be relaxed.

Figure 3 shows the initial CN definition derived from the sentence fragment "Unremarkable with the exception of mild shortness of breath and chronically swollen ankles." The domain expert has marked "shortness of breath" and "swollen ankles" with CN type "sign or symptom" and subtype "present". When BADGER analyzes this sentence, it assigns the complex noun phrase "the exception of mild shortness of breath and chronically swollen ankles" to a single prepositional phrase buffer. When a complex noun phrase has multiple head nouns or multiple modifiers, the class constraint becomes a conjunctive constraint. Class constraints on words such as "unremarkable" which are of class <Root Class> are dropped as vacuous.

```
CN-type: Sign or Symptom
Subtype: Present
Extract from Prep. Phrase "WITH"
Verb = <NULL>
Subject constraints:
    words include   "UNREMARKABLE"
Prep. Phrase constraints:
    preposition =   "WITH"
    words include
     "THE EXCEPTION OF MILD SHORTNESS OF
      BREATH AND CHRONICALLY SWOLLEN ANKLES"
    modifier class <Sign or Symptom>
    head class
       <Sign or Symptom>, <Body Location or Region>
```

Figure 3: An initial CN definition, with exact words from the instance

It is highly unlikely that this CN definition will ever apply to a sentence from a different text, but it is guaranteed to operate properly on the sentence that motivated it. The initial CN definitions created from a training corpus are too tightly constrained to be useful until CRYSTAL relaxes some of the constraints. Semantic constraints are relaxed by moving up the semantic hierarchy or by dropping the constraint. Exact word constraints are relaxed by dropping all but a subsequence of the words or dropping the constraint. The combinatorics on ways to relax constraints becomes overwhelming. There are over 57,000 possible generalizations of the initial CN definition in Figure 3.

The CRYSTAL algorithm:

```
Initialize Dictionary and Training Instances Database
Do until no more initial CN definitions in Dictionary
    D = an initial CN definition removed
        from the Dictionary
    Loop
        D' = the most similar CN definition to D
        If D' = NULL, exit loop
        U = the unification of D and D'
        Test the coverage of U in Training Instances
        If the error rate of U > Tolerance
            exit loop
        Delete all CN definitions covered by U
        Set D = U
    Add D to the Dictionary
Return the Dictionary
```

The following section shows how CRYSTAL induces generalizations from a set of initial CN definitions, testing that each proposed definition does not over generalize.

### 3.2 Inducing Generalized CN Definitions

CRYSTAL finds useful generalizations of its initial CN definitions by locating and comparing definitions that are highly similar. Let D be the definition being generalized. Assume we have found a second definition, D', which is very similar to D according to a similarity metric that counts the number of relaxations required to unify two CN definitions. A new definition U is then created with constraints relaxed just enough to unify D and D'. The new definition U is then tested against the training corpus to make sure that it does not extract phrases that were not marked with the CN type and subtype being learned.

If U is a valid CN definition, CRYSTAL deletes from the dictionary all definitions covered by U, thus reducing the size of the dictionary while still covering all the positive training instances. In particular, D and D' will be deleted. The definition U becomes the current definition and this process is repeated, using similar CN definitions to guide the further relaxation of constraints. Eventually a point is reached where further relaxation would produce a definition that exceeds some pre-specified error tolerance. At that point, CRYSTAL begins the generalization process on another initial CN definition until all initial definitions have been considered for generalization.

Faced with an exponential number of ways in which the constraints could be relaxed, CRYSTAL relaxes exactly those constraints that allow the current CN definition to be unified with a similar definition. An implementation that allows CRYSTAL to locate similar definitions efficiently is discussed in the next section.

CRYSTAL unifies two similar definitions by finding the most restrictive constraints that cover both. If word constraints from the two definitions have an intersecting string of words, the unified word constraint is that intersecting string. Otherwise the word constraint is dropped.

Unifying two class constraints may involve moving up the semantic hierarchy to find a common ancestor of classes in the two constraints. Class constraints are dropped entirely when they reach the root of the semantic hierarchy. If a constraint on a particular syntactic buffer is missing from one of the two definitions, that constraint is dropped from the unified constraints.

As an example of unifying two CN definitions, suppose that one definition has the class constraint <Sign or Symptom> for the subject buffer and the other has the class constraint <Laboratory or Test Result>. These unify to <Finding>, their common parent in the semantic hierarchy. If the direct object of one definition has a class constraint requiring both <Disease or Syndrome> and <Acquired Abnormality> and the other only requires <Disease or Syndrome>, then the unified class constraint on direct object will have only <Disease or Syndrome>.

Rather than reprocessing the training texts each time it tests the validity of a proposed CN definition, CRYSTAL uses the BADGER sentence analyzer to segment the training documents and creates a database of instances. This database includes an entry for each segment of the training corpus, not just those with phrases marked as having relevant information. If an instance meets all the constraints of a CN definition, but the phrase extracted had not been tagged with the appropriate CN type and subtype, it is counted as an error. CRYSTAL accommodates a limited amount of noise in the training data by using an error tolerance threshold instead of calling a definition bad from a single extraction error. This adds robustness to the dictionary, which is needed when dealing with unrestricted text.

The algorithm presented here has glossed over some implementation issues of how to find the most similar CN definitions or test the coverage and error rate of a proposed CN definition against the training instances efficiently. The following section discusses these issues, which can be critically important in scaling up to a large training corpus.

### 3.3 Efficiency Issues: Finessing Intractability

Since each CN definition may have several constraints and a variety of ways to relax each constraint, there are an exponential number of generalizations possible for a given CN definition. CRYSTAL has the challenge of producing a near optimal dictionary while avoiding intractability and maintaining a rich expressiveness of its CN definitions.

CRYSTAL reduces the intractable problem of constraint relaxation to the easier problem of finding a similar CN definition. Relaxing the constraints to unify a CN definition with a similar definition has the effect of retaining the constraints shared with another valid definition and dropping accidental features of the current definition. This is also guaranteed to produce a CN definition with greater coverage that either of the definitions being unified, since the coverage of the two CN definitions is disjoint.

Finding similar definitions efficiently is achieved by indexing the CN definitions database by verbs and by extraction buffers. In this way, CRYSTAL can retrieve a list of similar definitions which is small relative to the entire database. Each of these is tested with a similarity metric that looks for intersecting classes and intersecting strings of words for corresponding syntactic buffers.

Testing a generalized definition's error rate on the training corpus is actually done on the BADGER Instances Database, a database of instances which have already been segmented by the BADGER sentence analyzer. The primary index is on verbs, including the <null> verb for sentence fragments. Testing a CN definition that has a constraint on the verb can be done by retrieving a small percent of the instances, with most of the constraint testing done on pointers in memory without the need to retrieve the actual instance. CRYSTAL drops the constraint on exact verb only after relaxing all other constraints as far as possible, to take full advantage of the efficiency of indexing by verb.

CRYSTAL has been tested on a corpus of 385 hospital discharge reports, averaging just under one thousand words each, which produced 14,719 training instances with 2,122 positive instances of "diagnosis" and 6,047 positive instances of "sign or symptom". CRYSTAL is able to induce a dictionary of all CN types and subtypes from this training set in about 10 minutes of clock time on a DEC ALPHA AXP 3000 using 45 MB of memory.

## 4 Experimental Results

Experiments were conducted in which 385 annotated hospital discharge reports were partitioned into a training set and a blind test set. Dictionaries for each CN type and subtype were induced from the training set and then evaluated on the test set. Performance is measured here in terms of recall and precision, where recall is the percentage of possible phrases that the dictionary extracts and precision is the percentage correct of the extracted phrases. For example if there are 5,000 phrases that could possibly be extracted from the test set by a dictionary, but the dictionary extracts only 3,000 of them, recall is 60%. If the dictionary extracts 4,000 phrases, only 3,000 of them correct, precision is 75%.

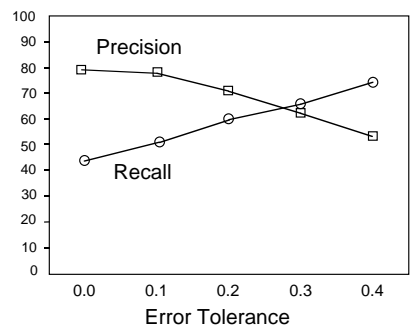

Figure 4: Effect of the error tolerance setting on performance

The choice of error tolerance parameter has a significant impact on performance and can be used to manipulate a tradeoff between recall and precision of a dictio-nary. Figure 4 shows performance of a dictionary of CN definitions that identify "sign or symptom" of any subtype, where the error tolerance is varied from 0.0 to 0.4. The results shown here are the averages of 50 random partitions of the corpus into 90% training and 10% test documents for each error tolerance.

There is another parameter that affects recall and precision. The results in Figure 4 are for generalized CN definitions, those with coverage >= 2. Precision can be boosted by raising the minimum coverage threshold. At error tolerance of 0.0 the dictionary for "sign or symptom" had precision 79 and recall 44 at minimum coverage of 2, precision 87 and recall 35 at minimum coverage of 5, and precision 91 and recall 24 at minimum coverage of 10.

The dictionary for all CN types and subtypes had 194 CN definitions that covered 10 or more training instances, 527 that covered from 3 to 9, and 793 with coverage of 2. This is for a dictionary induced at error tolerance 0.2 from all 14,719 training instances.

To assess the learning curve as training size increases, we set the training partition at 10%, 30%, 50%, 70%, and 90% of the 385 annotated documents. This was done 50 times for each training size and results averaged. The number of positive training instances in a partition depends on what CN type and subtype is being learned.

The graph in Figure 5 plots recall against the number of positive training instances for the most frequent CN type and subtypes in the corpus. Recall is over 60 and still increasing at this level of training for "diagnosis" and "sign or symptom" of any subtype. With this error tolerance set at .20 and minimum coverage at 2, precision remains fairly constant regardless of training size, at about 70 for most CN types and subtypes.

The difference in coverage for "symptom, absent" and "symptom, present" is due to a limitation in negation handling by the current version of CRYSTAL. CN definitions for absent symptoms can include a constraint requiring the word "no", but CRYSTAL has no mechanism to require the absence of "no" for present symptoms. The next version of CRYSTAL will handle conjunctions, disjunction, and the scoping of negation.

Here are a few representative CN definitions to give a feeling of what CRYSTAL is learning. Symptom, present is extracted from the direct object when the verb is "revealed" and the direct object has the word "a" and head class <Finding>. This covers cases where a test reveals "a mass", "a pleural effusion", "a murmur" and so forth. If the constraint on "a" is relaxed the CN definition would erroneously identify "no murmurs" or "normal bowel sounds" which are symptoms of subtype absent.

A similar CN definition that identifies symptom, absent in the direct object requires the verb "revealed" and the word "no" in the direct object. This has over 90% accuracy with no semantic constraints. Another CN definition with no semantic constraints finds symptom, absent in a prepositional phrase with the preposition "WITHOUT", covering 229 training instances with 13% error rate. If a constraint is added to require the class <Finding> in the prepositional phrase, coverage

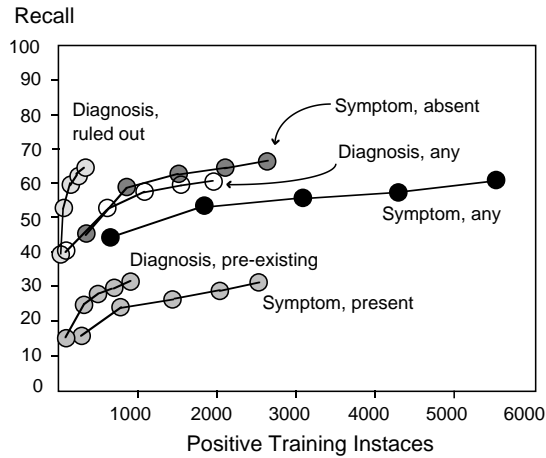

Figure 5: Learning curve: increase of coverage with training size

drops to 75 with 5% errors.

Some CN definitions look for specific words, such as "normal", "unremarkable", "regular", and "negative" to identify symptom, absent, or "enlarged", "increased", "mild", and "supple" to identify symptom, present.

The lack of semantic constraints on most of the high coverage CN definitions may be attributed to gaps in the semantic lexicon. We refrained from customizing the UMLS medical MetaThesaurus for our particular task, since our main research interest is designing portable systems with a minimum of knowledge engineering. UMLS, unfortunately, has spotty coverage of clinical terms. It's remarkable that we have gotten good performance from a lexicon that lacks words such as "lesions", "rate", "rhythm", "tenderness", and "distention".

## 5 Related Work

Previous research on inductive learning in natural language processing has concentrated on the semantics of isolated words [Granger, 1977; Mooney, 1987; Zernik, 1991]. CRYSTAL is one of the first systems to automatically induce a dictionary of information extraction rules. Only two of the seventeen research sites participating in the ARPA-sponsored Fifth Message Understanding Conference [MUC-5, 1993] described automatically generated dictionaries, the University of Massachusetts and the University of Southern California.

The UMass system used the Autoslog dictionary construction tool [Riloff, 1993], which generates a proposed CN definition for each phrase to be extracted from a motivating instance in the text. AutoSlog uses heuristics to select certain exact words from the instance as "trigger words", often selecting the head verb. The semantic constraints on the extracted buffer are set in advance by the user. AutoSlog made no attempt to relax constraints, merge similar CN definitions, or test proposed definitions on the training corpus. Each proposed definition had to be reviewed by a human who retained the definitions that looked reasonable, and discarded those that were more dubious (roughly 70% of AutoSlog's pro-posed definitions had to be manually discarded).

The PALKA system [Moldovan and Kim, 1992; Moldovan et al., 1993] used by USC includes an induction step similar to CRYSTAL. PALKA constructs an initial "Frame-Phrasal pattern structure" (FP-structure) from each example clause that has been marked as having information to be extracted. The FP-structure includes a constraint on the root form of the verb and semantic constraints on noun groups except for prepositional phrases with no marked information. PALKA generalizes the semantic constraints by moving up the semantic hierarchy and specializes by moving down until the FP-structure covers all the applicable positive training instances and none of the negative. PALKA is not tolerant of noise in the training data, and a single negative instance can block an otherwise good generalization.

CRYSTAL allows more expressive extraction patterns than either AutoSlog or PALKA. While AutoSlog requires an exact word constraint on the trigger word or words, which was determined heuristically from a single instance, CRYSTAL allows an exact word constraint on any words it learns to be significant, and it can also learn CN definitions with no word constraints. PALKA's FP-structures constrain the root form of the verb but allow no other exact word constraints. Unlike AutoSlog and PALKA, CRYSTAL makes no *a priori* decision on which constituents are to be included in its CN definitions.

The inductive concept learning in CRYSTAL is similar to an inductive learning algorithm described by Mitchell [1982], a "specific-to-general" data-driven search to find the most specific generalization that covers all positive and no negative instances. CRYSTAL has the same goal, but uses a greedy unification of similar instances rather than breadth-first search. This does not guarantee that CRYSTAL will always find the optimal dictionary, but in practice, CRYSTAL dictionaries appear to be near optimal.

Another survey of inductive concept learning, by Michalski [1983] describes a "star methodology" that is quite similar to CRYSTAL algorithm. This methodology is capable of learning a set of multiple generalizations to cover the positive instances by keeping only the best unification rather than branching on all possible generalizations. Michalski sees this methodology in terms of a set covering algorithm, with the goal of merging generalizations to find the minimum set that cover all positive training instances, while avoiding negative instances. This is exactly the goal of CRYSTAL.

## 6 Conclusions

CRYSTAL is one of the first systems to automatically derive a conceptual dictionary from a training corpus, and represents an improvement over previous attempts to derive text analysis rules from training examples. The goal of CRYSTAL is to find the minimum set of generalized CN definitions that cover all of the positive training instances and to test each proposed definition against the training corpus to ensure that the error rate is within a predefined tolerance. CRYSTAL's error tolerance parameter also allows a user to manipulate a recall-precision tradeoff.

The requirements of CRYSTAL are a sentence analyzer, a semantic lexicon that maps individual words into classes in a semantic hierarchy, and a set of annotated training texts. The approach taken by CRYSTAL facilitates a turnkey information extraction system that requires no familiarity with language processing technologies on the part of the end-user who creates an annotated training corpus. CRYSTAL then uses the training to build a fully functional conceptual dictionary that requires no further knowledge engineering.

## References

[Granger, 1977] R. Granger. FOUL-UP: A program that figures out meanings of words from context. In *Proceedings of the Fifth International Joint Conference on Artificial Intelligence*, pages 172–178. Morgan Kaufmann, 1977.

[Lehnert et al., 1993] W. Lehnert, J. McCarthy, S. Soderland, E. Riloff, C. Cardie, J. Peterson, F. Feng, C. Dolan, and S. Goldman. University of Massachusetts/Hughes: Description of the CIRCUS system as used for MUC-5. In *Proceedings of the Fifth Message Understanding Conference (MUC-5)*, pages 277–290, 1993.

[Lehnert, 1991] W. Lehnert. Symbolic/subsymbolic sentence analysis: Exploiting the best of two worlds. In J. Barnden and J. Pollack, editors, *Advances in Connectionist and Neural Computation Theory, Vol. 1*, pages 135–164. Ablex Publishers, Norwood, NJ, 1991.

[Lindberg et al., 1993] D. Lindberg, B. Humphreys, and A. McCray. Unified medical language systems. *Methods of Information in Medicine*, 32(4):281–291, 1993.

[Michalski, 1983] R. S. Michalski. A theory and methodology of inductive learning. *Artificial Intelligence*, 20:111–161, 1983.

[Mitchell, 1982] T. M. Mitchell. Generalization as search. *Artificial Intelligence*, 18:203–226, 1982.

[Moldovan and Kim, 1992] D. Moldovan and J. Kim. PALKA: A system for linguistic knowledge acquisition. Technical Report PKPL 92-8, USC Department of Electrical Engineering Systems, 1992.

[Moldovan et al., 1993] D. Moldovan, S. Cha, M. Chung, T. Gallippi, K. Hendrickson, J. Kim, C. Lin, and C. Lin. USC: Description of the SNAP system used for MUC-5. In *Proceedings of the Fifth Message Understanding Conference (MUC-5)*, pages 305–319, 1993.

[Mooney, 1987] R. Mooney. Integrated learning of words and their underlying concepts. In *Proceedings of the Ninth Annual Conference of the Cognitive Science Society*, pages 974–978, 1987.

[MUC-5, 1993] *Proceedings of the Fifth Message Understanding Conference (MUC-5)*, San Mateo, CA, August 1993. Morgan Kaufmann.

[Riloff, 1993] E. Riloff. Automatically constructing a dictionary for information extraction tasks. In *Proceedings of the Eleventh National Conference on Artificial Intelligence*, pages 811–816, Washington, DC, July 1993. AAAI Press / MIT Press.

[Zernik, 1991] U. Zernik. *Lexical Acquisition: Exploiting On-Line Resources to Build a Lexicon*. Lawrence Erlbaum Associates, Hillsdale, NJ, 1991.